\pdfoutput=1

\documentclass[11pt]{article}

\usepackage[dvipsnames]{xcolor}
\usepackage[final]{acl}
\usepackage{soul}
\usepackage{float}
\usepackage{times}
\usepackage{latexsym}
\usepackage{adjustbox}
\usepackage[T1]{fontenc}
\usepackage{enumitem}
\usepackage{tabularx}
\usepackage[utf8]{inputenc}
\usepackage[most]{tcolorbox}

\usepackage{microtype}

\usepackage{inconsolata}

\usepackage{graphicx}
\usepackage{listings}
\title{SummExecEdit: A Factual Consistency Benchmark in Summarization with Executable Edits}

\author{
  Onkar Thorat \quad Philippe Laban \quad Chien-Sheng Wu \\
  Salesforce AI Research \\
  \texttt{\{othorat, wu.jason\}@salesforce.com}
}

\newcommand{\dataname}{SummExecEdit }

\begin{document}
\maketitle
\begin{abstract}
Detecting factual inconsistencies in summarization is critical, yet existing benchmarks lack the necessary challenge and interpretability for robust evaluation. In this paper, we introduce SummExecEdit, a novel pipeline and benchmark leveraging executable edits to assess models on their ability to both detect factual errors and provide accurate explanations. The top-performing model, Claude3-Opus, achieves a joint detection and explanation score of only 0.49 in our benchmark, with individual scores of 0.67 for detection and 0.73 for explanation. We conduct detailed evaluations to assess the current state of models in this field and find that more than half of the 20+ LLMs in our study struggle with over 30\% of the \dataname benchmark.
Additionally, we identify four primary types of explanation errors, with 45.4\% of them involving a focus on completely unrelated parts of the summary.

\end{abstract}

\section{Introduction}

Large language models (LLMs) have shown remarkable performance across various tasks by generating coherent responses while facing a major challenge with factual hallucination \cite{acm-llm-hallucination}. Various evaluation methods \cite{laban-summac, alex-summeval, fabbri2022qafactevalimprovedqabasedfactual, tang2023understandingfactualerrorssummarization, luo2023chatgptfactualinconsistencyevaluator, yang2024sifidreassesssummaryfactual} and inconsistency detection benchmarks have been proposed, which used approaches such as
editing ground truth texts to create intentional inconsistencies \cite{laban-etal-2023-summedits}, or generating benchmarks with LLMs where they label the factual accuracy of target texts post-generation \cite{tang2024tofuevalevaluatinghallucinationsllms}.

However, previous approaches often involve broad and sweeping edits, i.e., trivial and multiple edits within a single sample \cite{laban-etal-2023-summedits}. \citet{kim2024fablesevaluatingfaithfulnesscontent} relied entirely on humans to generate edits in their work because initial experiments with LLM-based edits primarily produced trivial changes.
LLM-generated multiple trivial edits make factual inconsistency errors easier for LLMs to detect. For instance, GPT-4 achieves an 82.4\% detection accuracy on the SummEdits inconsistency benchmark \cite{laban-etal-2023-summedits}. However, relying on humans to generate edits and benchmarks is neither scalable nor cost-efficient. Therefore, we need mechanisms to generate more complex edits in a scalable and cost-effective manner.
Moreover, high-quality explanations for inconsistencies in these detection benchmarks are missing. Such explanations are essential for evaluating a model's ability to justify its detection results and reason over factual inconsistencies.

To address these drawbacks, we propose to leverage the concept of executable edits \cite{laban2024beyond} to generate a challenging and interpretable benchmark called \dataname - extending SummEdits~\footnote{Our data is released at \url{https://huggingface.co/datasets/Salesforce/summexecedit}}. 
While normal editing rewrites the entire summary with edits already applied, executable editing focuses on isolating and replacing a substring in the original summary, allowing precise changes that can introduce factual inconsistencies. 
Such edits helps LLM to better concentrate and come up with a complex yet granular and controlled edit with a meaningful explanation for the same in a structured way, helping to eliminate broad and sweeping multiple edits. 

\begin{figure*}
    \centering
    \includegraphics[width=\textwidth]{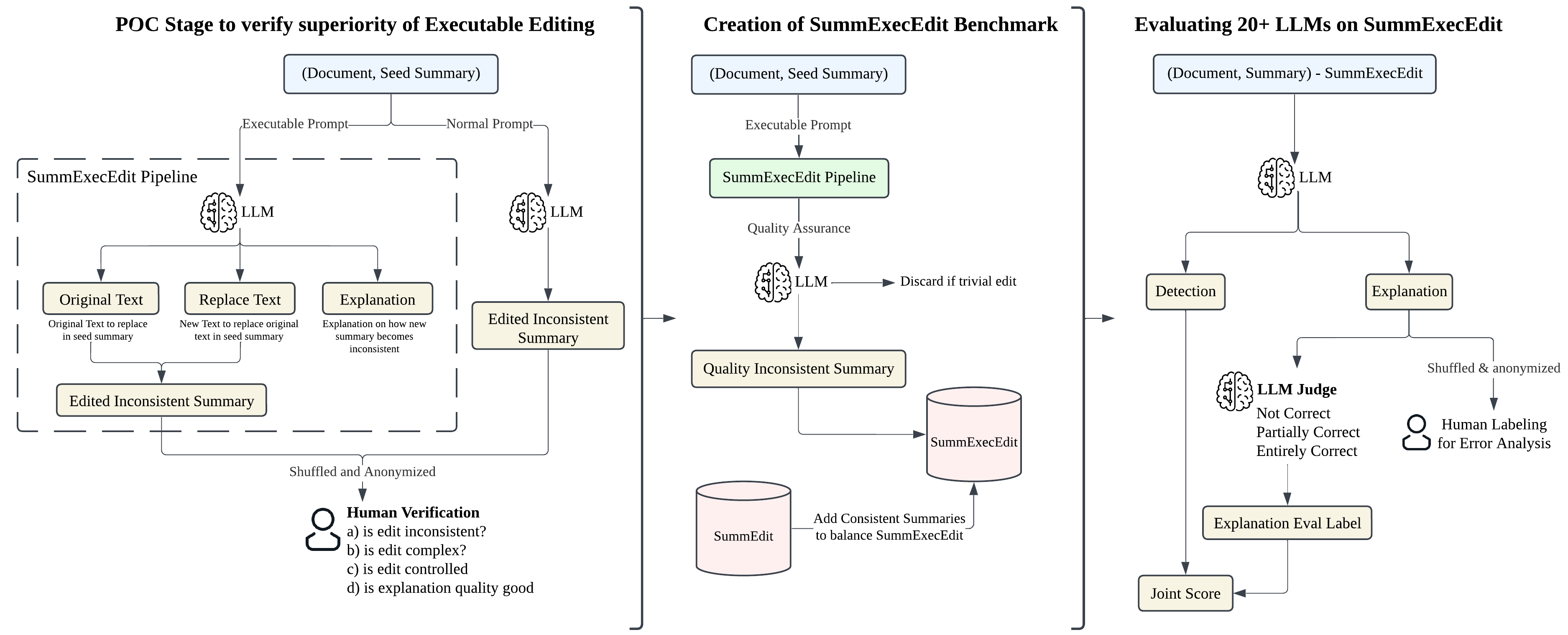}
    \caption{Overall contribution of our work. We start by first verifying and proving the superiority of executable editing over directly editing a summary. We use human annotation to label four different parameters. We then use SummExecEdit pipeline to create the full benchmark ensuring quality is not hampered when we scale up from our POC. We finally evaluate variety of LLMs on this benchmark using two different prompt types. For simplicity, we have not shown the minor details. Pipelines like SummExecEdit can be further used for creation of such datasets and benchmarks.}
    \label{fig:contri}
\end{figure*}

\begin{figure}[h!]
\centering
\resizebox{0.45\textwidth}{!}{
\begin{tcolorbox}[
    title=\textbf{Example of SummExecEdit}, colframe=SpringGreen,
    colback=white, 
    width=\textwidth,
    sharp corners=south, 
    enhanced,
    boxrule=0pt, 
    coltitle=white, colbacktitle=SpringGreen, fonttitle=\bfseries, 
]

\textbf{\textcolor{black}{Original Summary}}  
\begin{tcolorbox}[colback=LimeGreen!3, colframe=SpringGreen, sharp corners, boxrule=0.5pt]
The document depicts the banishment of Coriolanus, deemed an enemy to the people and his country. Despite the sadness of his family and friends, Coriolanus departs, promising to make them proud and maintain contact while they hope for a chance to repeal his banishment.
\end{tcolorbox}

\textbf{\textcolor{black}{Generating inconsistent summary with SummExecEdit}}
\begin{tcolorbox}[colback=white, colframe=SpringGreen, sharp corners, boxrule=0.5pt]
\textbf{\textcolor{black}{Text to Replace}}  
\begin{tcolorbox}[colback=LimeGreen!3, colframe=SpringGreen, sharp corners, boxrule=0.5pt]
\textcolor{red!80}{"deemed an enemy to the people and his country"}
\end{tcolorbox}

\textbf{\textcolor{black}{Replaced Text}}  
\begin{tcolorbox}[colback=LimeGreen!3, colframe=SpringGreen, sharp corners, boxrule=0.5pt]
\textcolor{OliveGreen}{"deemed a traitor to the people and his country"}
\end{tcolorbox}

\textbf{\textcolor{black}{Edited Summary generated with help of Executable Prompt}}  
\begin{tcolorbox}[colback=LimeGreen!3, colframe=SpringGreen, sharp corners, boxrule=0.5pt]
The document depicts the banishment of Coriolanus, \textcolor{red}{\st{deemed an enemy to the people and his country}} \textcolor{OliveGreen}{deemed a traitor to the people and his country}. Despite the sadness of his family and friends, Coriolanus departs, promising to make them proud and maintain contact while they hope for a chance to repeal his banishment.
\end{tcolorbox}

\textbf{\textcolor{black}{Explanation}}  
\begin{tcolorbox}[colback=LimeGreen!3, colframe=SpringGreen, sharp corners, boxrule=0.5pt]
The document states that Coriolanus is deemed an enemy, but the summary escalates this to calling him a traitor, which has a more severe connotation not supported by the document.
\end{tcolorbox}
\end{tcolorbox}
\textbf{\textcolor{black}{Inconsistent summary from SummEdits benchmark}}  
\begin{tcolorbox}[colback=LimeGreen!3, colframe=SpringGreen, sharp corners, boxrule=0.5pt]
The document depicts the banishment of Coriolanus, deemed an enemy to his people \textcolor{red}{\st{and his country}} \textcolor{OliveGreen}{but a friend to his country}. Despite the sadness of his family and \textcolor{red}{\st{friends}} \textcolor{OliveGreen}{foes}, Coriolanus departs, promising to make them proud and \textcolor{red}{\st{maintain contact}} \textcolor{OliveGreen}{sever contact} while they hope for a chance to repeal his banishment.
\end{tcolorbox}

\end{tcolorbox}
}

\caption{An example of comparison of inconsistent summaries. For SummExecEdit summary, LLM generates the text that should be replaced in the original summary with an alternative inconsistent text and an explanation for the inconsistency. The edit is controlled and granular - difficult to identify but still meaningful enough. An inconsistent summary directly generated with a direct prompt makes multiple trivial edits that are easy to find.}
\label{fig-example}
\end{figure}


Through our work, we aim to introduce an executable editing style for creating benchmarks and datasets, present a new benchmark - \dataname for evaluating LLMs in hallucination detection and explanation within real-world systems, and thus identify the current challenges LLMs face in this field.
We compare executable edits to normal direct editing and show that executable edits are superior, providing in range of 18-25\% of higher number of challenging samples with better explanations. 
We evaluate a wide range of open-source and open-api access LLMs on \dataname and find that most of these models lack on the combined task of detection and explanation of error, with the best model - Claude3-Opus achieving a joint score of only 0.49, with individual scores of 0.67 for detection and 0.73 for explanation. We find 60  
inconsistent summaries being detected as consistent by all 20+ LLMs in our study and analyze their categories.
We also perform error analyses for incorrect explanations generated by candidate LLMs and categorize them into four reasons, with 45.4\% of errors focusing on completely unrelated parts of the summary. Overall contribution of our work is as shown in Figure \ref{fig:contri}.



\begin{table*}[t]
    \centering
    \resizebox{\textwidth}{!}{
    \begin{tabular}{lccccc}
        \hline
        \textbf{Condition} & \textbf{N} & \textbf{\%Controlled ($\uparrow$)} & \textbf{\%Inconsistent ($\uparrow$)} & \textbf{\%Complex ($\uparrow$)} & \textbf{\%Explanation ($\uparrow$)}\\
        \hline
        GPT4-Turbo (Exec) & 144 & 86.81 & 81.25 & 45.83 & 44.44\\ 
        GPT4-Turbo (Non-Exec) & 134 & 90.30 & 81.34 & 23.13 & 20.90\\
        GPT3.5-Turbo (Exec) & 134 & 76.87 & 73.13 & 17.16 & 16.42\\
        GPT3.5-Turbo (Non-Exec) & 133 & 86.47 & 72.18 & 18.05 & 12.03\\
        Claude3-Opus (Exec) & 138 & 92.03 & 84.78 & \textbf{49.28} & \textbf{48.55}\\
        Claude3-Opus (Non-Exec) & 136 & \textbf{97.06} & \textbf{88.24} & 31.62 & 30.15\\
        \hline
    \end{tabular}
    }
    \caption{Manually annotated scores for comparison between executable and non-executable edits. Each column shows percentage of N but does not consider the edits filtered out in its previous columns.}
    \label{tab:execvsnonexec}
\end{table*}

\section{Related Works}

\textbf{Methods employed in creating previous benchmarks.} Several previous works have assessed various metrics in relation to human judgments associated with hallucination and factual inconsistencies. Works prior to advances in language models focused on human and expert level \cite{alex-summeval, falke-etal-2019-ranking} while the very recent works focused on generating annotations with the help of language models \cite{chhikara2024lamsumcreatingextractivesummaries, tang2024tofuevalevaluatinghallucinationsllms, laban-etal-2023-summedits}. Although these models generate coherent texts with a good level of inter-annotator agreements with respect to consistency \cite{laban-etal-2023-summedits}, such benchmarks may struggle to keep the changes in the original data to minimal \cite{nguyen-etal-2024-llms}. Executable edits with help of language models was first explored by \citet{laban2024beyond}, where  it focused on replacing a substring for document editing. We build upon the works of SummEdits \cite{laban-etal-2023-summedits} to use executable edits to generate new factual consistency benchmark.\newline
\textbf{Detecting Factual Inconsistencies.} Since automatic evaluation metrics such as ROUGE \cite{lin2004rouge} or BLEU \cite{papineni2002bleu} cannot reliably detect factual inconsistencies, several non-LLM based methods have been developed such as question-based \cite{fabbri2022qafactevalimprovedqabasedfactual} and entailment-based \cite{laban-summac, zha-etal-2023-alignscore, tang2024minicheckefficientfactcheckingllms}. \citet{gekhman2023trueteacherlearningfactualconsistency} introduced TrueTeacher - a method for generating synthetic data by annotating diverse model-generated summaries using a LLM that can enhance the performance of a state-of-the-art model. In various recent works, LLMs have shown superior performance when used as a judge to evaluate factual consistencies \cite{luo2023chatgptfactualinconsistencyevaluator}. We explore this further in various settings and also re-evaluate modern-day LLMs on our benchmark generated using executable edits.

\section{Executable Edits}
\label{section:exec_edits}

Given a \textit{(source, target)} pair with a goal to edit the target such that it becomes factually inconsistent with the source, we can either ask the LLM to rewrite the target token-by-token, or ask it to isolate and highlight only a specific substring in the target text to be replaced. 
The latter can also be considered a program that can be executed and used. 
Executable editing is simple but it can help us generate more controlled and complex edits. 

Moreover, this minimizes synthetic data in the benchmarks since a majority of the original data remains same. An example of SummExecEdit is given in Figure \ref{fig-example}


We first experiment and verify on a smaller sample to show that executable edits work better compared to non-executable edits. We selected around 100 original (document, summary) pairs from \citet{laban-etal-2023-summedits} and used GPT4-Turbo, Claude3 Opus, and GPT3.5-Turbo to generate edited summaries and explanation of inconsistencies using both executable and non-executable prompts (Appendix \ref{prompt}) in a structured json format.
This generates around 600 edits, which are shuffled, anonymized, and annotated by two of the authors manually based on the four questions: 
a) is the edit inconsistent? 
b) is the edit complex/good quality? 
c) is the edit controlled/granular? 
d) is the explanation quality good?
Each annotator annotated nearly 400 edits with 200 in common to verify the inter-annotator agreement in Table \ref{tab:exec-iaa}.

The result for our manual annotation is given in Table \ref{tab:execvsnonexec}. We use a filtering mechanism - each subsequent column filters out the edits deemed inappropriate by either of the annotators in the previous column.
The models show a similar trend - executable edits lead to a higher score towards the end implying a higher number of good edits and explanations. 
For example, Claude3-Opus provides nearly 18\% more controlled and high quality edits with executable prompt.

\section{\dataname Benchmark}
We leverage \citet{laban-etal-2023-summedits} to build our benchmark across 10 domains such as News, Podcast, Bills, Sales calls, etc. 
Based on Table \ref{tab:execvsnonexec}, we use both GPT4-Turbo and Claude3-Opus models to generate our benchmark using the executable prompt. Both the models are asked to generate six edits for each (document, summary) pair.

To ensure we maintain quality control, after generating these edits, we remove the trivial edits with help of GPT4-Turbo. We ask GPT4-Turbo to classify an edit as a date change, number change, antonym change, or others. Any edit classified as date, number, or antonym change is removed from the benchmark. The prompt used in here is mentioned \ref{appendix-remove-trivial-prompt}.

The final benchmark results in 
2,121 factually inconsistent summaries. To balance out the inconsistent summaries in the benchmark, we use 2,120 factually consistent edits from SummEdits benchmark resulting in a total of 4,241 samples in the final \dataname benchmark. Each of the 10 domains provide around 200-300 inconsistent summaries.
The distribution of each domains is given in Table \ref{tab:benchmark-stats}. 

\begin{table}[h!]
    \centering
    \begin{tabular}{lcc}
        \hline
        \textbf{Domain} & \textbf{N} & \textbf{\%Inconsistent}\\
        \hline
        SciTLDR & 307 & 58.31\\ 
        News & 567 & 47.8\\
        Podcast & 344 & 58.72\\
        BillSum & 608 & 44.9\\
        SamSum & 450 & 51.11\\
        Shakespeare & 511 & 41.3\\
        QMSum & 334 & 50.3\\
        ECTSum & 416 & 47.12\\
        Sales Email & 368 & 57.1\\
        Sales Call & 336 & 53.87\\
        \hline
        \textbf{Total} & 4241 & 50 \\
        \hline
    \end{tabular}
    \caption{Statistics of the domains in \dataname}
    \label{tab:benchmark-stats}
\end{table}

\section{Results}

To evaluate LLMs on \dataname benchmark, we use two types of prompts. \textbf{D\&E} - Detect and Explain error. Models need to detect if there is any factual inconsistency in summary, if so, explain the inconsistency (Appendix \ref{appendix-promptv1}). \textbf{E|D} - Explain error given Detection. Given that the summary is inconsistent, models need to explain the inconsistency in the summary (Appendix \ref{appendix-promptv2}).

Table \ref{tab:final_scores} provides the Detection Accuracy (DA) of all the models using prompt D\&E. 
The best performing model on \dataname, Claude3.5-Sonnet, provides an accuracy of only around 73\%, showing challenging nature of the benchmark. 

\begin{table*}[h!]
    \centering
        \begin{tabular}{lcccc|cccc}
        & \multicolumn{4}{c|}{\textbf{Correlation}} & \multicolumn{4}{c}{\textbf{BAcc}}\\
        \hline
        \textbf{Model} & \textbf{V1} & \textbf{V2} & \textbf{V3} & \textbf{V4} & \textbf{V1} & \textbf{V2} & \textbf{V3} & \textbf{V4}\\
        \hline
            claude3-haiku & 0.076 & 0.478 & 0.732 & 0.725 & 0.385 & 0.49 & 0.655 & 0.618\\
            claude3-opus & 0.484 & 0.782 & 0.801 & 0.774 & 0.46 & 0.76 & 0.776 & 0.738\\
            gpt3.5-turbo & 0.103 & 0.118 & 0.325 & 0.563 & 0.375 & 0.401 & 0.495 & 0.625\\
            gpt4o & 0.603 & 0.816 & 0.804 & \textbf{0.833} & 0.533 & 0.792 & 0.794 & \textbf{0.811}\\
        \hline
        \end{tabular}
        \caption{Correlation and Balanced Accuracy of prompts and models for explanation evaluation with respect to manual evaluation.}
    \label{tab:explanation_relations}
\end{table*}

\subsection{Analysis for Detections}
The overall detection results show that many LLMs struggle in detecting the factual error.
As a reference, we also evaluate two non-LLM based approaches which use far lesser compute - AlignScore \cite{zha-etal-2023-alignscore} receiving 57.4\% and MiniCheck \cite{tang2024minicheckefficientfactcheckingllms} receiving 60.0\% accuracy, which are better than only two open-source LLMs.

We analyze our findings and observe that 11 or more LLMs incorrectly detected nearly 1,300 out of 4,241 samples, indicating that over half of the modern day LLMs in our study struggle with more than 30\% of our benchmark. Additionally, we find that 65 samples were misclassified by all LLMs. Among these, 60 samples belonged to the category of inconsistent summaries that LLMs incorrectly identified as consistent, highlighting their difficulty in detecting minor inconsistencies. Upon manually reviewing these 60 inconsistent summaries, we find that the errors primarily fall into three categories - \textbf{Ungrounded information} - summary contains information not grounded in the document, \textbf{Reasoning Error} - summary makes wrong inferences from information in the document, and \textbf{Nuanced Meaning Shift} - subtle shift in meaning between the document and the summary. This shows that LLMs struggle most at such cases.
Examples of such misclassified cases are provided in Appendix \ref{appendix-examples}


\begin{table}[H]
    \centering
    \resizebox{\linewidth}{!}{
        \begin{tabular}{lcccc|c}
        \hline
        & \multicolumn{4}{c|}{\textbf{D\&E}} & \textbf{E|D}\\
        \hline
        \textbf{Model} & DA($\uparrow$) & DS($\uparrow$) & ES($\uparrow$) & JS($\uparrow$) & ES($\uparrow$)\\
        \hline
            claude3-opus & 0.71 & 0.67 & 0.733 & \textbf{0.491} & 0.684\\
            gemini-1.5-pro & 0.728 & 0.728 & 0.665 & 0.484 & 0.628\\
            gpt-4o & 0.733 & 0.725 & 0.629 & 0.456 & 0.684\\
            gpt4-turbo & 0.729 & 0.581 & 0.714 & 0.415 & \textbf{0.711}\\
            gemini-2.0-flash & 0.69 & 0.431 & 0.893 & 0.386 & 0.651\\
            claude3.5-sonnet & \textbf{0.737} & 0.537 & 0.686 & 0.369 & 0.706\\
            gemini-1.5-flash & 0.708 & 0.556 & 0.633 & 0.352 & 0.62\\
            gpt3.5-turbo & 0.585 & 0.676 & 0.467 & 0.315 & 0.488\\
            gemini-2.0-flash-lite & 0.652 & 0.345 & 0.812 & 0.28 & 0.638\\
            command-r-plus & 0.655 & 0.369 & 0.703 & 0.259 & 0.595\\
            gemini-pro & 0.625 & 0.432 & 0.499 & 0.216 & 0.456\\
            claude3-haiku & 0.622 & 0.54 & 0.387 & 0.209 & 0.473\\
            claude3-sonnet & 0.636 & 0.33 & 0.626 & 0.206 & 0.633\\
            command-r & 0.609 & 0.414 & 0.489 & 0.202 & 0.49\\
            palm-bison & 0.608 & 0.248 & 0.473 & 0.117 & 0.407\\
            \hline
            mixtral-8x7b & 0.609 & 0.73 & 0.503 & 0.367 & 0.568\\
            mistral-large & 0.703 & 0.512 & 0.714 & 0.366 & 0.624\\
            llama3.1-70b & 0.711 & 0.52 & 0.747 & 0.388 & 0.58\\
            llama3-70b & 0.678 & 0.423 & 0.713 & 0.302 & 0.585\\
            llama3.1-8b & 0.537 & 0.872 & 0.309 & 0.27 & 0.478\\
            llama3-8b & 0.555 & 0.73 & 0.239 & 0.175 & 0.458\\
            mistral-7b & 0.541 & 0.107 & 0.735 & 0.079 & 0.488\\
        \hline
        \end{tabular}
        }
        \caption{Scores for D\&E (Prompt Detect and Explain error) and E|D (Prompt Explain error given Detection) for inconsistent summaries. DA - Detection Accuracy, DS - Detection Score, ES - Explanation Score, JS - Joint Score. ES evaluated by EvalV4 prompt using GPT4o. DA based on entire benchmark. DS, ES, and JS based only on inconsistent summaries. Last 7 models are open-source.}
    \label{tab:final_scores}
\end{table}

It is worth analyzing whether newer models improve upon their predecessors in detection performance. While GPT-4o outperforms GPT-4 Turbo in detection accuracy, it struggles with explaining its reasoning, scoring lower than GPT-4 Turbo. In contrast, Claude 3.5 Sonnet and Meta's Llama3.1 80b shows a clear improvement over Claude 3 Sonnet and Llama3 80b respectively across all aspects. A major concern arises with the Gemini Flash models. Although Gemini 2.0 Flash gets better at the explanation quality, it exhibits lower detection accuracy compared to Gemini 1.5 Flash. This raises a question if new models are truly getting better at detecting hallucination and other general purpose classification tasks. Llama3.1-8b model improves upon previous Llama3-8b but it seems that the model is more biased towards saying that the summary is inconsistent depicted from its lower detection accuracy of 0.537 but comparatively higher detection score of 0.872. Such a phenomenon is also observed for many other models.

\subsection{Evaluating Explanations}
Two of the authors manually annotated around 1200 explanations with 300 in common. The results for manual annotation are given in appendix in Table \ref{tab:explanation_manual_scores}.
Based on these annotations we evaluated different LLM-as-Judge \cite{llm-as-judge} by asking them to assign a label as Entirely Correct (1), Partially Correct (0.5), or Not Correct (0).
We use following four types of prompts for explanation evaluation using GPT4o, GPT3.5-Turbo, Claude3-Opus, and Claude3-Haiku.
\begin{itemize}[leftmargin=*]
    \item Reference less and no seed (EvalV1) - We provide document, edited summary, and LLM's explanation to be evaluated.
    \item Reference less and with seed (EvalV2) - We provide seed summary (ground truth), edited summary, and LLM's explanation to be evaluated.
    \item Reference based with seed (EvalV3) - We provide seed summary (ground truth), edited summary, reference explanation, and LLM's explanation to be evaluated.
    \item Reference only (EvalV4) - We provide reference explanation and LLM's explanation to be evaluated.
\end{itemize}

The correlations for four prompts with different LLMs with respect to our manually annotated explanations are provided in Table \ref{tab:explanation_relations}. Selected explanations for manual annotations were shuffled and randomly selected. The model and either of the two prompts that generated those explanations were anonymized. The IAA between both the annotators are - Correlation of 0.885 and Cohen Kappa of 0.81. We think that the reference explanations generated at the time of edit itself are the best and of high quality, scoring 0.95 for 40 samples. The prompt EvalV4 works the best which suggests that evaluating explanations or reasoning of models, works better when we have access to what the edit/reason is or a reference explanation. 
Thus, good and challenging benchmarks need to also provide high quality explanations.

\subsection{Joint Scores on \dataname}
We define the joint scores comprising of both detection and explanation scores on factually inconsistent summaries in \dataname. 
\textbf{Detection Score (DS)} is calculated only on 2,121 factually inconsistent summaries, and we 
give a score of 1 if model correctly detects the summary being factually incorrect and 0 otherwise.
\textbf{Joint score (JS)} is calculated by multiplying both DS and ES element-wise, and the results are presented in Table \ref{tab:final_scores}. 

The best model Claude3-Opus achieves a JS of 0.49 which suggests that the task of detecting factual inconsistency and explaining the same is still a challenging task for most modern-day LLMs, making them incapable to reason out-of-the-box.
It is worth noting the big JS gap for open-api and open-source models.
At the same time, it is also good to see Mixtral-8x7b achieving the best DS. This also brings up an interesting finding - some models are good at detecting the factual errors but struggle to explain the error, and also vice-versa. 
Models belonging to the same family also show differing behavior, for example GPT-4o and GPT4-Turbo or Mixtral-8x7b and Mistral-Large show different trends.

\subsection{Error Analysis for Explanations}
We analysed 350 of our manually annotated explanations that were incorrect or partially correct and observed that most of the errors in these explanations mainly fall under the following categories. If an explanation contains multiple errors, we report the first found error in the explanation.\newline
\textbf{Misattribution of Error} - This is the most common type of error, accounting for 45.4\% of incorrect explanations. The explanation would focus on a completely unrelated part of the summary or the document and assign the blame on it.\newline
\textbf{Additional Unrelevant Explanation} - The LLM provides the correct explanation but also continues to generate some unrelated explanation. Such explanations make 28.9\% of incorrect explanations.\newline
\textbf{Concentrating on Completeness} - The explanation focuses on completeness showing missing details in summary rather than focusing on factual correctness. This accounts for 15.4\% of incorrect explanations.\newline
\textbf{Vague Explanation} - These are either complex to understand or incomplete explanations missing out on details. They may correctly identify the error but not effectively explain it. They account for 10.3\% of incorrect explanation.

While we do not find any relation of errors with specific models or specific prompts, we find a relation where different models happen to make similar errors in explanations belonging to the same document and factually inconsistent summary pairs.


\section{Conclusion}
In this work, we first explored the executable editing with LLMs for generating benchmarks. Through our experiments, we show its superiority in generating a challenging benchmark and use it to generate a new factual inconsistency detection and explanation benchmark - \dataname. We evaluate various LLMs over this benchmark for both detection and explanation of factual inconsistencies. The best performing model for detection - Claude3.7-Sonnet achieves an accuracy of 73\% while Claude3-Opus performs best on combined task of detection and explanation achieving a joint score of 0.49. These scores suggest that \dataname is a challenging benchmark. We explore various prompt formats that lead to better auto-evaluation by LLMs and perform error-analysis for incorrect explanations generated by LLMs. We present detection and explanation analysis that can help LLM Developers for evaluating LLMs’ abilities to detect factual errors and reason about facts before releasing the models or deploying in systems. We also hope researchers and practitioners can now use executable edits to generate datasets and benchmarks.

\section*{Limitations}
While executable edits help in creating more challenging benchmark, it is important to note a few limitations of these edits and this work in general. First, generating benchmarks with executable edits require availability of ground truth \textit{(source, target)} pairs which might not always be the case. Second, non-executable edits are equally capable of generating controlled edits but these are simple and low quality. If the end-goal is to only generate controlled edits and no-complex edits, non-executable edits could be used as well. Last, we have only experimented using these edits for creating a single benchmark related to summarization. Further experiments in other domains are required.



\bibliography{custom}

\appendix

\section{Model Details}
\textbf{OpenAI Models.} We used three OpenAI models in our experiments - GPT4o, GPT4-Turbo, and GPT3.5-Turbo. All the models were accessed from the official OpenAI API.
\newline
\textbf{Google Models.} We used six Google models in our experiments - Gemini-1.5-Pro, Gemini-1.5-Flash, Gemini-Pro, Gemini-2.0 Flash, Gemini-2.0 Flash-Lite, and Palm-Bison. We used Google Cloud Platform to access Google models, which relies on the VertexAI API.
\newline
\textbf{Anthropic, Cohere, Llama3, Mistral models.} We used the Claude models in Anthropic family, Command-R models in Cohere family, models in Llama3 family, and models in Mistral family on AWS bedrock.

\section{License Details and Terms of Use}
We leverage \citet{laban-etal-2023-summedits} for raw data which was released under Apache-2.0 license, granting copyright license.



\section{Manual Annotation and Inter-Annotator Agreement}

In Section \ref{section:exec_edits}, we manually annotate the generated inconsistent summaries to compare between executable and non-executable edits based on following 4 questions:\newline
\textbf{Is the edit inconsistent?} - We annotate if the edited summary is factually inconsistent or the replaced text still kept the edited summary consistent.\newline
\textbf{Is the edit complex/good quality?} - We annotate if the factual inconsistency is complex or interesting, meaning the edits should not be trivial such as easy antonym swap, negating sentence, or numerical and date changes. Trivial edits make the benchmark easier.\newline
\textbf{Is the edit controlled/granular?} - We annotate if the edit is controlled and granular or still very broad and sweeping, a drawback of previous benchmarks. \newline
\textbf{Is the explanation quality good?} - We annotate if the reference explanation specifying the reason of inconsistency generated by LLM at the time of edit is correct.

Table \ref{tab:exec-iaa} provides the Cohen Kappa as the inter-annotator agreement.
\begin{table}[h!]
    \centering
    \begin{tabular}{lcc}
        \hline
        \textbf{Question} & \textbf{N} & \textbf{Cohen Kappa}\\
        \hline
        Inconsistent & 202 & 0.76\\ 
        Interesting & 177 & 0.61\\
        Controlled & 156 & 0.6\\
        Explanation quality & 43 & 0.49\\
        \hline
    \end{tabular}
    \caption{Inter-annotator agreement between both the authors. We keep filtering out the examples in every row to remove the bias of non-matching annotations since the subsequent annotation of questions for every example depend on the previous questions. The Cohen for Explanation looks low but the disagreement is on a very small samples. Since Cohen Kappa is looking at classes with equal weights and due to a big imbalance, it has a larger effect.}
    \label{tab:exec-iaa}
\end{table}

Table \ref{tab:explanation_manual_scores} provides explanation scores for our manual annotations of explanation evaluation. We can see a comparable performance gap between the open-source and open-api access models.
\begin{table}[]
    \centering
        \begin{tabular}{lcc|cc}
        \hline
        & \multicolumn{2}{c|}{\textbf{Prompt V1}} & \multicolumn{2}{c}{\textbf{Prompt V2}}\\
        \hline
        \textbf{Model} & N & ES($\uparrow$) & N & ES($\uparrow$)\\
        \hline
            claude3-opus & 29 & \textbf{0.897} & 39 & \textbf{0.884} \\
            gpt-4o & 28 & 0.821 & 39 & \textbf{0.884} \\
            gpt4-turbo & 26 & 0.807 & 20 & 0.8\\
            mixtral-8x7b & 31 & 0.516 & 40 & 0.662\\
            gpt3.5-turbo & 30 & 0.567 & 40 & 0.7\\
            llama3-70b & 20 & 0.775 & 40 & 0.725\\
            command-r-plus & 17 & 0.706 & 40 & 0.762\\
            gemini-pro & 17 & 0.676 & 40 & 0.587\\
            claude3-haiku & 25 & 0.54 & 40 & 0.537\\
            claude3-sonnet & 19 & 0.763 & 40 & 0.725\\
            command-r & 15 & 0.6 & 40 & 0.625\\
            llama3-8b & 36 & 0.319 & 40 & 0.625\\
            palm-bison & 10 & 0.6 & 40 & 0.6\\
            mistral-7b & 6 & 0.833 & 40 & 0.612\\
        \hline
        \end{tabular}
        \caption{Scores for prompt V1 and V2 for inconsistent summaries. N - Number of samples, ES - Explanation Score. ES for Manual Annotation based on 887 explanations annotated by one author. The score for 40 manually annotated reference explanation is 0.95.}
    \label{tab:explanation_manual_scores}
\end{table}

\section{\dataname Examples}
\label{appendix-examples}

\begin{figure}[h!]
\resizebox{0.48\textwidth}{!}{
\begin{tcolorbox}[
    title=\textbf{Example 1}, 
    colframe=black, colback=white, 
    width=\textwidth,
    sharp corners=south,
    enhanced,
    boxrule=0.8pt,
    coltitle=white, colbacktitle=black, fonttitle=\bfseries,
]

\textbf{\textcolor{black}{Original Summary:}}  
\begin{tcolorbox}[colback=gray!10, colframe=gray!50, sharp corners, boxrule=0.5pt]
A conversation between a sales agent and a potential client discussing different types of insurance products, including home, auto, life, and health insurance. The sales agent provides information on different home insurance plans and pricing, as well as available discounts for clients with good credit scores and other factors.
\end{tcolorbox}

\textbf{\textcolor{black}{Text to Replace:}}  
\begin{tcolorbox}[colback=gray!5, colframe=gray!40, sharp corners, boxrule=0.5pt]
"A conversation between a sales agent and a potential client"
\end{tcolorbox}

\textbf{\textcolor{black}{Replaced Text:}}  
\begin{tcolorbox}[colback=gray!5, colframe=gray!40, sharp corners, boxrule=0.5pt]
"A conversation between a sales agent and a new client"
\end{tcolorbox}

\textbf{\textcolor{black}{Edited Summary:}}  
\begin{tcolorbox}[colback=gray!10, colframe=gray!50, sharp corners, boxrule=0.5pt]
A conversation between a sales agent and a new client discussing different types of insurance products, including home, auto, life, and health insurance. The sales agent provides information on different home insurance plans and pricing, as well as available discounts for clients with good credit scores and other factors.
\end{tcolorbox}

\textbf{\textcolor{black}{Explanation:}}  
\begin{tcolorbox}[colback=gray!10, colframe=gray!50, sharp corners, boxrule=0.5pt]
The document indicates this is a potential client, but the summary refers to them as a new client, which is not supported.
\end{tcolorbox}
\end{tcolorbox}
}
\caption{Appendix D - Example 1 - SummExecEdit}
\end{figure}

\begin{figure}[h!]
\resizebox{0.48\textwidth}{!}{
\begin{tcolorbox}[
    title=\textbf{Example 2}, 
    colframe=black, colback=white, 
    width=\textwidth,
    sharp corners=south,
    enhanced,
    boxrule=0.8pt,
    coltitle=white, colbacktitle=black, fonttitle=\bfseries,
]

\textbf{\textcolor{black}{Original Summary:}}  
\begin{tcolorbox}[colback=gray!10, colframe=gray!50, sharp corners, boxrule=0.5pt]
Abraham Lincoln Bicentennial 1-Cent Coin Redesign Act - Directs the Secretary of the Treasury, during 2009, to issue one-cent coins with the reverse side bearing four different designs representing different aspects of the life of Abraham Lincoln.
\end{tcolorbox}

\textbf{\textcolor{black}{Text to Replace:}}  
\begin{tcolorbox}[colback=gray!5, colframe=gray!40, sharp corners, boxrule=0.5pt]
"during 2009"
\end{tcolorbox}

\textbf{\textcolor{black}{Replaced Text:}}  
\begin{tcolorbox}[colback=gray!5, colframe=gray!40, sharp corners, boxrule=0.5pt]
"starting in 2009"
\end{tcolorbox}

\textbf{\textcolor{black}{Edited Summary:}}  
\begin{tcolorbox}[colback=gray!10, colframe=gray!50, sharp corners, boxrule=0.5pt]
Abraham Lincoln Bicentennial 1-Cent Coin Redesign Act - Directs the Secretary of the Treasury, starting in 2009, to issue one-cent coins with the reverse side bearing four different designs representing different aspects of the life of Abraham Lincoln.
\end{tcolorbox}

\textbf{\textcolor{black}{Explanation:}}  
\begin{tcolorbox}[colback=gray!10, colframe=gray!50, sharp corners, boxrule=0.5pt]
The document specifies that the coins are to be issued during 2009, but the summary implies they will be issued starting from 2009, potentially extending beyond that year.
\end{tcolorbox}
\end{tcolorbox}
}
\caption{Appendix D - Example 2 - SummExecEdit}
\end{figure}

\begin{figure}[h!]
\resizebox{0.48\textwidth}{!}{
\begin{tcolorbox}[
    title=\textbf{Example 3}, 
    colframe=black, colback=white, 
    width=\textwidth,
    sharp corners=south,
    enhanced,
    boxrule=0.8pt,
    coltitle=white, colbacktitle=black, fonttitle=\bfseries,
]

\textbf{\textcolor{black}{Original Summary:}}  
\begin{tcolorbox}[colback=gray!5, colframe=gray!40, sharp corners, boxrule=0.5pt]
Oman is opening up its airspace to Israeli carriers, including El Al. Previously, the airlines had to fly a roundabout route to parts of Asia, adding hours to journey times.
\end{tcolorbox}

\textbf{\textcolor{black}{Text to Replace:}}  
\begin{tcolorbox}[colback=gray!5, colframe=gray!40, sharp corners, boxrule=0.5pt]
"Oman is opening up its airspace to Israeli carriers, including El Al."
\end{tcolorbox}

\textbf{\textcolor{black}{Replaced Text:}}  
\begin{tcolorbox}[colback=gray!5, colframe=gray!40, sharp corners, boxrule=0.5pt]
"Oman has opened up its airspace to Israeli carriers, including El Al."
\end{tcolorbox}

\textbf{\textcolor{black}{Edited Summary:}}  
\begin{tcolorbox}[colback=gray!10, colframe=gray!50, sharp corners, boxrule=0.5pt]
Oman has opened up its airspace to Israeli carriers, including El Al. Previously, the airlines had to fly a roundabout route to parts of Asia, adding hours to journey times.
\end{tcolorbox}

\textbf{\textcolor{black}{Explanation:}}  
\begin{tcolorbox}[colback=gray!10, colframe=gray!50, sharp corners, boxrule=0.5pt]
The document states that Oman announced it will open its airspace, implying a future event, but the summary says Oman has already opened it, suggesting it has already happened.
\end{tcolorbox}
\end{tcolorbox}
}
\caption{Appendix D - Example 3 - SummExecEdit}
\end{figure}

\begin{figure}[h!]
\resizebox{0.48\textwidth}{!}{
\begin{tcolorbox}[
    title=\textbf{Example 4}, 
    colframe=black, colback=white, 
    width=\textwidth,
    sharp corners=south,
    enhanced,
    boxrule=0.8pt,
    coltitle=white, colbacktitle=black, fonttitle=\bfseries,
]

\textbf{\textcolor{black}{Original Summary:}}  
\begin{tcolorbox}[colback=gray!5, colframe=gray!40, sharp corners, boxrule=0.5pt]
Harry and Jacob discuss their preparedness for an exam that will comprise a combination of multiple choice and open question essay formats, and how knowing the format of the test can influence their study strategy.
\end{tcolorbox}

\textbf{\textcolor{black}{Text to Replace:}}  
\begin{tcolorbox}[colback=gray!5, colframe=gray!40, sharp corners, boxrule=0.5pt]
"knowing the format of the test can influence their study strategy."
\end{tcolorbox}

\textbf{\textcolor{black}{Replaced Text:}}  
\begin{tcolorbox}[colback=gray!5, colframe=gray!40, sharp corners, boxrule=0.5pt]
"knowing the format of the test is crucial for developing their study strategy."
\end{tcolorbox}

\textbf{\textcolor{black}{Edited Summary:}}  
\begin{tcolorbox}[colback=gray!10, colframe=gray!50, sharp corners, boxrule=0.5pt]
Harry and Jacob discuss their preparedness for an exam that will comprise a combination of multiple choice and open question essay formats, and how knowing the format of the test is crucial for developing their study strategy.
\end{tcolorbox}

\textbf{\textcolor{black}{Explanation:}}  
\begin{tcolorbox}[colback=gray!10, colframe=gray!50, sharp corners, boxrule=0.5pt]
While the document suggests that the test format can influence study strategy, the summary exaggerates this by stating that knowing the format is crucial for developing the strategy.
\end{tcolorbox}
\end{tcolorbox}
}
\caption{Appendix D - Example 4 - SummExecEdit}
\end{figure}

\begin{figure}[h!]
\resizebox{0.48\textwidth}{!}{
\begin{tcolorbox}[
    title=\textbf{Example 5}, 
    colframe=black, colback=white, 
    width=\textwidth,
    sharp corners=south,
    enhanced,
    boxrule=0.8pt,
    coltitle=white, colbacktitle=black, fonttitle=\bfseries,
]

\textbf{\textcolor{black}{Original Summary:}}  
\begin{tcolorbox}[colback=gray!5, colframe=gray!40, sharp corners, boxrule=0.5pt]
The speaker is discussing the topic of nutrition, sharing their experience with following the Zone diet and recommending it as a way to learn about food and develop a better relationship with it. They express frustration with nutrition narratives that vilify certain foods or diets, and instead encourage focusing on the benefits of a particular diet, without putting down other approaches.
\end{tcolorbox}

\textbf{\textcolor{black}{Text to Replace:}}  
\begin{tcolorbox}[colback=gray!5, colframe=gray!40, sharp corners, boxrule=0.5pt]
"and instead encourage focusing on the benefits of a particular diet, without putting down other approaches."
\end{tcolorbox}

\textbf{\textcolor{black}{Replaced Text:}}  
\begin{tcolorbox}[colback=gray!5, colframe=gray!40, sharp corners, boxrule=0.5pt]
"and instead encourage focusing on the benefits of a plant-based diet, without putting down other approaches."
\end{tcolorbox}

\textbf{\textcolor{black}{Edited Summary:}}  
\begin{tcolorbox}[colback=gray!10, colframe=gray!50, sharp corners, boxrule=0.5pt]
The speaker is discussing the topic of nutrition, sharing their experience with following the Zone diet and recommending it as a way to learn about food and develop a better relationship with it. They express frustration with nutrition narratives that vilify certain foods or diets, and instead encourage focusing on the benefits of a plant-based diet, without putting down other approaches.
\end{tcolorbox}

\textbf{\textcolor{black}{Explanation:}}  
\begin{tcolorbox}[colback=gray!10, colframe=gray!50, sharp corners, boxrule=0.5pt]
The document mentions the speaker wants people to focus on the benefits of their chosen diet in general, but the summary specifies a plant-based diet, which is not stated in the document.
\end{tcolorbox}
\end{tcolorbox}
}
\caption{Appendix D - Example 5 - SummExecEdit}
\end{figure}

\section{Annotation Tool}
\begin{figure*}[b!]
    \centering
    \includegraphics[width=\linewidth]{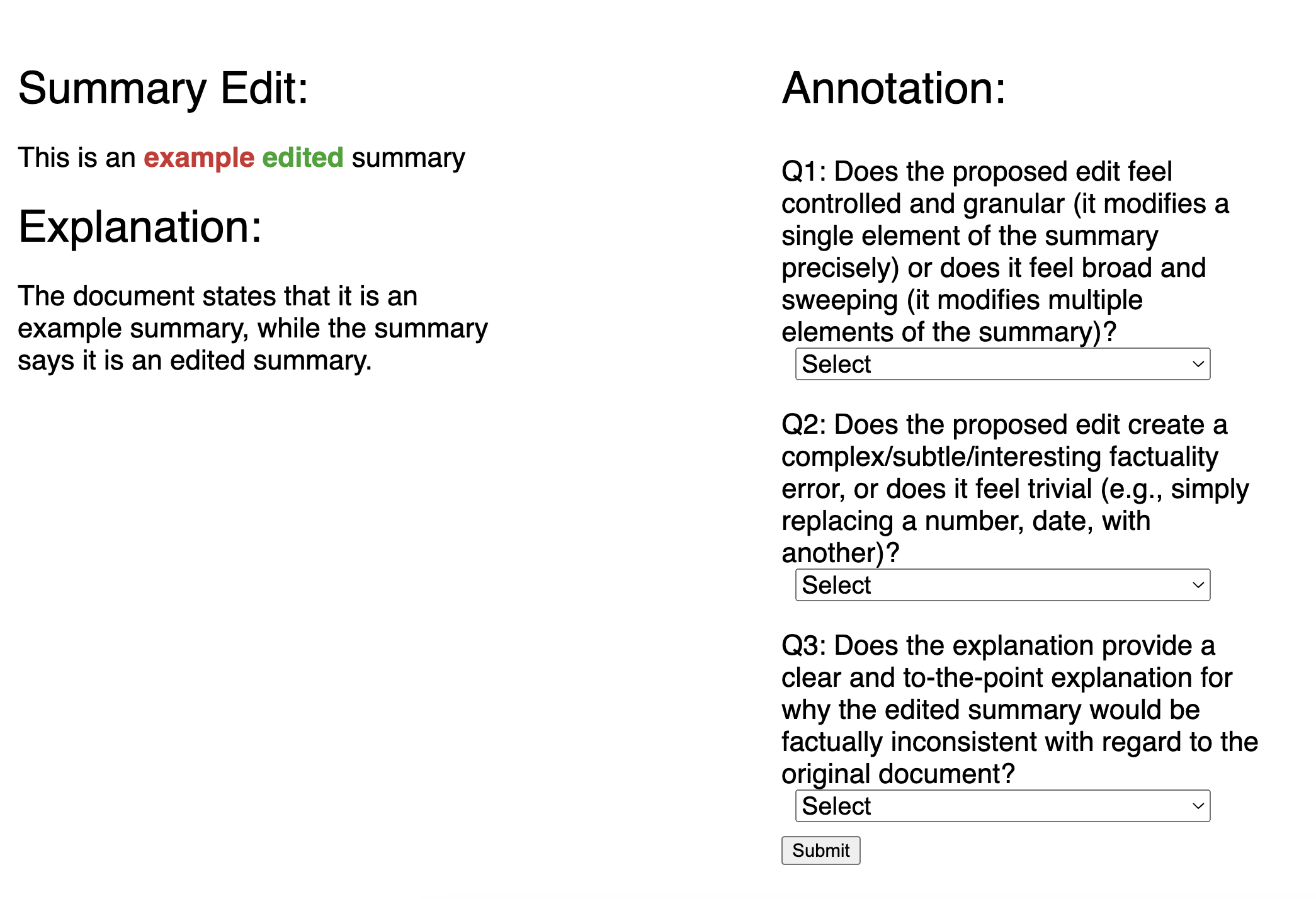}
    \caption{Two-Column annotation interface with highlights used for annotation}
    \label{fig:annot-tool}
\end{figure*}

Figure \ref{fig:annot-tool} provides an example of our annotation tool. The text highlighted in red, is replaced by the text highlighted in green and on the right side we perform the required annotation. This helps us annotate more reliably and easily.

\onecolumn
\section{Prompts}
\label{prompt}

\subsection{Executable Prompt}
\label{appendix-exec-prompt}
\begin{lstlisting}
You are given a document and its corresponding summary. You must generate 6 edits to the summary which should strictly follow the guidelines mentioned below. 

While making edits, think of what in a summary could be changed, added, or removed that is difficult to detect but makes the summary inconsistent with the document. Be mindful, that you cannot change the figures, numbers, dates, digits mentioned in the summary.

The edit should introduce an error which makes the summary factually incorrect, or inconsistent with the document. Few examples of good error types are given after the guidelines. Refer to those edits and generate similar error types. For each edit only one of the error type should be used.

[Guidelines] 
- You can generate only one error per edited you generate.
- Your output should be a valid JSON string: that starts with : {"edits": [ ... ]} and contains edits in the format described below.
- Remember to not introduce following types of errors:
    - You cannot use antonyms or opposite words to introduce an error.
    - You cannot negate the sentence to introduce an error.
    - You cannot replace a date by some other random date or random time of the year to introduce an error.
    - You cannot replace some amount or money by other random amount or money.
    - You cannot replace a number or a percentage by some other random number or percentage to introduce an error.
    - You cannot replace a name by some other random name to introduce an error. 
- Your explanation should not mention the original summary or the edit, it should only describe the difference between the new summary and document with respect to the document: "The document says X but the document says Y."
- Make sure your "original_text" is a substring of the summary.
- You should diversify the types of errors you generate as much as possible. Do not generate all edits of the same type.
- In the explanations you generated, you must refer to the edited summary simply as the summary. You must not mention edits or editing, or the "edited summary" in any way, simply call it the summary.

[Edit format]
Each edit should be a valid JSON object, with three keys as show: "original_text", "replace_text", and "explanation". The "original_text" key should be the original text in the summary that you are editing which is a substring of the summary. The "replace_text" key should be the new text that you are replacing the original text with. The "explanation" key should be a 1-2 sentence explanation of why the edit is an error.

[Document]
[DOCUMENT]

[Summary]
[SUMMARY]
\end{lstlisting}

\subsection{Non-Executable/Normal Prompt}
\label{appendix-nonexec-prompt}
\begin{lstlisting}
You are given a document and its corresponding summary. You must generate 6 edits to the summary which should strictly follow the guidelines mentioned below. 

While making edits, think of what in a summary could be changed, added, or removed that is difficult to detect but makes the summary inconsistent with the document. Be mindful, that you cannot change the figures, numbers, dates, digits mentioned in the summary.

The edit should introduce an error that makes the summary factually incorrect, or inconsistent with the document.

[Guidelines] 
- You can generate only one error per edited you generate.
- You must not rewrite an entirely new summary. Your edited summary should modify, insert or delete AT MOST 5 words of the original summary. 
- Remember to not introduce following types of errors:
    - You cannot use antonyms or opposite words to introduce an error.
    - You cannot negate the sentence to introduce an error.
    - You cannot replace a date by some other random date or random time of the year to introduce an error.
    - You cannot replace some amount or money by other random amount or money.
    - You cannot replace a number or a percentage by some other random number or percentage to introduce an error.
    - You cannot replace a name by some other random name to introduce an error. 
- You should diversify the types of errors you generate as much as possible. Do not generate all edits of the same type.
- You do not need to mention the type of edit that you made.
- For each of the three edits, rewrite the entire summary with the edit you make.
- In the explanations you generated, you must refer to the edited summary simply as the summary. You must not mention edits or editing, or the "edited summary" in any way, simply call it the summary.

[Edit format]
You should generate six edits to the original summary, and for each generate an one-sentence explanation of what the error in the edited summary is.
Each edit should follow following json format:
{"edits": [
	{"edited_summary": "[Edited version of the summary]", "explanation": "A natural language explanation of the error."},
    ...
]}

[Document]
[DOCUMENT]

[Summary]
[SUMMARY]
\end{lstlisting}
 
\subsection{Detect and Explain error (D\&E)}
\label{appendix-promptv1}
\begin{lstlisting}
You are given a document and its corresponding summary which may or may not be factually correct and consistent with the document. Your task is to generate a valid json string (use escape characters if double quotes and new lines are used within value) that has following two fields:
1. "consistent" - This field gives whether the summary is factually correct and consistent with respect to the document. It should be "yes" if the summary is factually consistent with respect to the document and "no" otherwise.
2. "explanation" - If you set "consistent" to "no", then this field should give the explanation as why you think it is incorrect or inconsistent. If you set "consistent" to "yes", then the explanation should be an empty string. Example format:

{"consistent": "no", "explanation": "[...]"}
or
{"consistent": "yes", "explanation": ""}

Now complete the task for the following document, summary pair:

[Document]
[DOCUMENT]

[Summary]
[SUMMARY]
\end{lstlisting}

\subsection{Explain error given Detection (E|D)}
\label{appendix-promptv2}
\begin{lstlisting}
You are given a document and its corresponding summary. We know that the summary contains a factual error which renders it inconsistent with the facts in the document. Your task is to provide a 1-2 sentence explanation that identifies what fact or facts in the summary is/are inconsistent with the document. You should output your explanation in a valid json format (use escape characters for double quotes and new lines if used within value), in the following format:

{"explanation": "[...]"}

Now complete the task for the following document, summary pair:

[Document]
[DOCUMENT]

[Summary]
[SUMMARY]
\end{lstlisting}

\subsection{Remove Trivial Edits}
\label{appendix-remove-trivial-prompt}
\begin{lstlisting}
You are given an original text, a replace text that has some change with respect to the original text, and an explanation describing the change. Analyze all three and categorize the replace text into one of the following categories described below.

[Categories]
Category - DATE_CHANGE
Explanation - This is a factual error where a date, day, year, month, or time from original summary is changed with some other random date, day, year, month, or time in the edited summary.

Category - NUMBER_CHANGE
Explanation - This is a factual error where a number, percentage, dollar amount from original summary is changed with some other random number, percentage, or money amount in the edited summary.

Category - ANTONYM_CHANGE
Explantion - This is a factual error where the edited summary uses an antonym for a word in original summary or simply has introduced the word 'not' to make the original sentence opposite to the edited sentence.

Category - OTHER
Explanation - All other types of factual errors that do not fall under any of the above three categories.

You should return your answer in form of a json having the field as 'category' and corresponding category from the above mentioned ones.

[Original Text]
[OG_TEXT]

[Replace Text]
[REP_TEXT]

[Explanation]
[EXPLAINATION]

\end{lstlisting}

\end{document}